\documentclass[letterpaper]{article} 
\usepackage{aaai25}  
\usepackage{times}  
\usepackage{helvet}  
\usepackage{courier}  
\usepackage[hyphens]{url}  
\usepackage{graphicx} 
\usepackage{natbib}  
\usepackage{caption} 
\usepackage{algorithm}
\usepackage{algorithmic}

\usepackage{newfloat}
\usepackage{listings}
\usepackage{csquotes}
\DeclareCaptionStyle{ruled}{labelfont=normalfont,labelsep=colon,strut=off} 
\floatstyle{ruled}
\newfloat{listing}{tb}{lst}{}
\floatname{listing}{Listing}
\title{Dialogic Learning in Child-Robot Interaction: A Hybrid Approach to Personalized Educational Content Generation}
\author {
    Elena Malnatsky\textsuperscript{\rm 1}, 
    Shenghui Wang\textsuperscript{\rm 2}, 
    Koen V. Hindriks\textsuperscript{\rm 1}, 
    Mike E.U. Ligthart\textsuperscript{\rm 1}
}
\affiliations {
    \textsuperscript{\rm 1}Vrije Universiteit Amsterdam, Amsterdam, The Netherlands\\
    \textsuperscript{\rm 2}University of Twente, Enschede, The Netherlands\\
    e.malnatsky@vu.nl, shenghui.wang@utwente.nl, k.v.hindriks@vu.nl, m.e.u.ligthart@vu.nl
}

\begin{document}

\maketitle

\begin{abstract}
Dialogic learning fosters motivation and deeper understanding in education through purposeful and structured dialogues. Foundational models offer a transformative potential for child-robot interactions, enabling the design of personalized, engaging, and scalable interactions. However, their integration into educational contexts presents challenges in terms of ensuring age-appropriate and safe content and alignment with pedagogical goals. We introduce a hybrid approach to designing personalized educational dialogues in child-robot interactions. By combining rule-based systems with LLMs for selective offline content generation and human validation, the framework ensures educational quality and developmental appropriateness. We illustrate this approach through a project aimed at enhancing reading motivation, in which a robot facilitated book-related dialogues. 
\end{abstract}

\section{Introduction}

\textit{Dialogic learning}, an educational approach that emphasizes learning through structured, purposeful dialogues, has long been valued for its ability to foster critical thinking, deeper understanding and motivation in educational contexts \cite{flecha_sharing_2000, wells_dialogic_1999}.  

Strategies in such dialogues include open-ended questions to encourage deeper thinking, reciprocity for shared meaning-making, contextual relevance by linking materials to students' lives, and reflection \cite{alexander_towards_2008, racionero_dialogic_2010}. However, facilitating such personalized interactions in modern classrooms remains challenging, as fully individualized teaching is often unfeasible.

Social robots, powered by recent advances in foundational AI, such as large language models (LLMs),  hold transformative potential for creating purposeful, engaging, and highly personalized child-robot interactions at scale \cite{zhang_large_2023}. Beyond generic, high-level customization (e.g., for skill level), these systems could help compose dialogues personalized to the unique preferences, interests, and developmental needs of each child, as well as the specific content of the educational topic being discussed.  

However, while large language models (LLMs) excel at generating personalized and engaging content at scale, their limitations pose significant challenges in educational contexts. (1) Unpredictable Adherence to Structure: LLMs struggle to reliably adhere to predefined conversation scripts, often deviating from structured sequences designed for specific pedagogical goals. This unpredictability risks undermining intended learning outcomes. (2) Inconsistent Accuracy and Appropriateness: LLM outputs often lack guarantees of factual accuracy, developmental appropriateness, or alignment with educational objectives \cite{hadi_large_2023, pozdniakov_large_2024}.

In child-robot interactions (CRIs), where educational goals and child welfare are paramount, we argue that humans must retain control over both the social and educational structure of the interaction. Ensuring that all content is safe, age-appropriate, and purposeful is imperative. Therefore, we advocate for a "\textit{better safe than sorry}" stance. Meaning, leveraging LLMs while minimizing the risks of granting them excessive freedom.

In this paper, we propose a \textit{hybrid framework} for designing CRIs that implement dialogic learning. Our approach combines a rule-based system with selectively deployed, offline LLM-driven content generation validated through AI-based validation and human moderation. We are identifying two complementary perspectives on hybridness: (1) the collaborative synergy between human experts and AI when designing and validating educational content, and (2) the integration of distinct AI technologies (e.g., rule-based systems with LLMs) to balance personalization at scale while maintaining content appropriateness and pedagogical quality.

We further illustrate a real-life application of our hybrid method through the Robot Bookworm project, a large-scale deployment of dialogic learning driven, personalized book discussions in primary schools.

In summary, our contributions are fourfold. First, we introduce design principles for educational CRIs that implement dialogic learning.  Second, we propose a hybrid system design that combines a rule-based structure with LLM-driven personalization for scalable dialogue-based CRIs. Third, we offer real-world evidence from the Robot Bookworm project, highlighting the feasibility and impact of our approach in primary schools. Finally, we outline remaining challenges and future directions for responsible and scalable educational CRIs.

\section{Foundations of Dialogic Learning}

Dialogic learning, rooted in Socratic traditions \cite{benson_socratic_2011, delic_socratic_2016} and grounded in socio-cultural theories (Vygotsky, Bakhtin),  is an educational approach that fosters knowledge co-construction through structured, purposeful conversations \cite{vygotsky_mind_1978,bakhtin_speech_1986}. It teaches to engage critically, articulate personal perspectives, and explore diverse viewpoints \cite{flecha_sharing_2000, wells_dialogic_1999} through egalitarian dialogue \cite{habermas_reason_1984} rather than lecture or tutoring. 

Strategies in dialogic learning include using open-ended questions  (e.g., “What would you do differently?”) 
 to encourage deeper thinking rather than simple factual recall and rely on reciprocity rather than one-sided questioning. Contextual relevance is created by linking materials to learners’ lives and experiences, making learning more meaningful. Additionally, reflection on personal beliefs and experiences is encouraged through prompts (e.g., "How does this remind you of…?”) to foster self-awareness. Exploring diverse perspectives is another key element that develops empathy and critical thinking. Finally, scaffolding techniques are employed to support learners as they engage with complex ideas and material \cite{bakhtin_speech_1986, flecha_sharing_2000, alexander_towards_2008, racionero_dialogic_2010}.

The principles of dialogic learning can be implemented through various methods tailored to specific educational context. For instance, \textit{dialogic reading} \cite{morgan_dialogic_2008, whitehurst_picture_1994} demonstrates how book-related conversations can be transformed into interactive learning experiences  by using strategies like PEER (Prompt, Evaluate, Expand, and Repeat)  and CROWD (Completion, Recall, Open-ended, Wh-, and Distancing) \cite{morgan_dialogic_2008, justice_promising_2003, whitehurst_picture_1994}.

Implementing dialogic learning through social robots offers a promising opportunity to enhance education at scale. However, translating dialogic principles into effective child-robot interactions poses many challenges. In this paper, we focus on the challenge of designing scalable, educationally structured, personalized interactions guided by the core principles of dialogic learning, while prioritizing reliability and safety of content.

\section{Implementing Dialogic Learning in CRI}

\subsection{Educational Conversations in CRI}
Educational conversations in CRI have shown promise in enhancing children’s engagement, motivation, and learning outcomes across various domains, such as reading comprehension and STEM education \cite{ligthart_design_2023, belpaeme_social_2018, ligthart_memory-based_2022,michaelis_supporting_2019, konijn_use_2020}. While many existing studies integrate elements of dialogic learning, they are typically blended within other educational methods rather than forming the primary framework. In this study, we focus specifically on designing interactions where dialogic learning is the foundation.

Despite the potential of educational child-robot interactions, current implementations often fall short of delivering the depth and personalization required to truly support meaningful learning experiences. Existing systems tend to rely on rigid, rule-based dialogue frameworks, which, while consistent, often feel mechanical and fail to adapt to the unique interests or needs of individual learners. Advances in generative AI present an opportunity to address these shortcomings by enabling scalable, personalized interactions.

\subsection{Design Principles}

To facilitate effective dialogic learning, interactions must adhere to several foundational design principles. These principles are derived from the literature on dialogic learning and CRI. First, \textbf{education-centered design} is essential; interactions must align with specific educational objectives and be designed by pedagogic domain experts, translating these goals into actionable and well-defined elements. Second, \textbf{knowledge} plays a central role, requiring robots to have access to sufficient representation of the educational material relevant to the context of interaction. \textbf{Memory} is necessary for continuity and personalization, enabling robots to retain and utilize pertinent information. \textbf{Personalization} is a key component, ensuring that dialogues align with a child’s unique interests, preferences, developmental stage, and the educational content under discussion. \textbf{Engagement and enjoyment} play a vital role, particularly for long-term motivation; interactions should captivate and sustain the child’s interest. Finally, \textbf{safety and reliability} are critical, with content that is suitable and aligned with ethical standards. 

Meeting these requirements at scale, while balancing personalization and quality, demands an approach that is both innovative and cautious. We therefore propose a \textbf{hybrid framework} as a step towards responsibly leveraging LLMs in educational CRIs.

\subsection{Hybrid Approach}
The proposed hybrid framework integrates human expertise, rule-based systems, and LLMs to create scalable, personalized, and safe educational interactions. 

\textbf{1. Co-Design:} Human domain experts retain full control over defining objectives and designing the educational structure, while AI tools (e.g., LLMs) assist in creative ideation through hybrid collaboration. 

\textbf{2. Static Rule-based scaffolding}:  The entire backbone of interactions is rigid and rule-based, including predefined dialogue blocks for socially constructive purposes and user model learning interactions. 

\textbf{3. Dynamic LLM-Generated Content}: Specific dialogue blocks (e.g., questions, observations, humor), include placeholders populated with LLM-driven content generated offline. This allows personalization to user profiles, book summaries, and educational objectives while ensuring prudent use of LLMs, limiting their role to areas where manual content creation would be impractical.

3\textbf{. Validation pipeline}: LLM-generated content is created offline and undergoes validation through LLM-based evaluation, followed by final human moderation, before being incorporated into interactions.

\subsection{Case Study: Robot-Bookworm Project}
\begin{figure}
    \centering
    \includegraphics[width=1\linewidth]{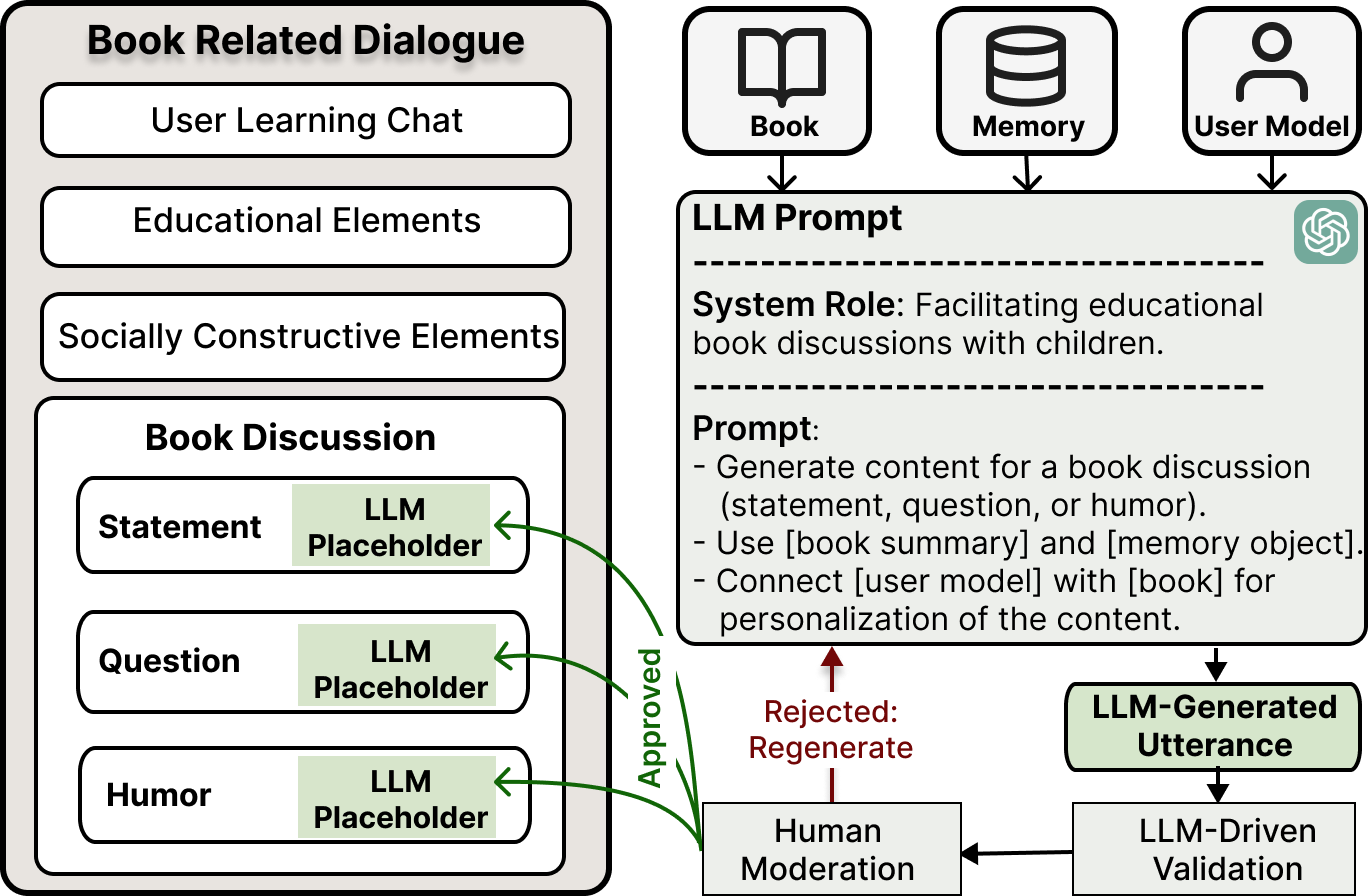}
    \caption{Hybrid Framework for Personalized Book-Related Dialogues in Child-Robot Educational Interactions}
    \label{fig:figure_1}
\end{figure}

The proposed hybrid framework was developed and evaluated in the real-world context of the Robot-Bookworm Project. This project is introduced in this paper to illustrate the practical application of the hybrid approach in a concrete use case. To this end, we provide an overview to contextualize and explain the framework; a dedicated paper will present the full details of the study and findings.

The educational goal of the Robot-Bookworm project was to enhance reading motivation among primary school children (ages 9–11) by facilitating personalized book-related dialogues through a social robot. Over the course of four weekly sessions, Leo the Nao robot engaged in book discussions with 100 children. Each child was uniquely assigned a different book, reading independently while discussing the book contents with Leo weekly. Drawing from pedagogic literature and insights from participatory design, alongside LLM-assisted ideation, we designed a structured educational framework. This framework incorporated evidence-based methods to foster reading motivation, with book-related dialogues as the central component, adhering to the principles outlined earlier. Dialogues were designed to foster a sense of relatedness to the book’s content, encouraging children to relate to the plot and reflect, and ultimately enhance reading enjoyment.

Generating over 1000 personalized utterances necessary to complete the book-related dialogic script across four sessions for 100 children, with access only to limited metadata from 100 distinct books, posed a significant challenge. To address this, we developed a hybrid framework (see Figure \ref{fig:figure_1}), which combined rule-based structures with selectively generated, offline-validated LLM content.
 
The backbone of the system was a predefined rule-based interaction script, ensuring consistency and educational quality. This structure was carefully designed by human interaction designers to align with the educational goals. LLMs were employed selectively in scenarios where rule-based personalization would be resource-intensive, such as generating unique dialogues for 100 children with individual books (see Figure \ref{fig:figure_1}, right side).

The interaction design adhered to the principles of the hybrid approach to designing dialogic learning outlined earlier:
\\\textbf{1. Rule-Based Scaffolding}: These static, rule-based components were fully predefined by human designers to ensure consistency, alignment with educational goals, and developmental appropriateness. They did not rely on LLM-generated content and included the following:
    \begin{itemize}
        \item \textbf{Socially Constructive Elements}: Fixed conversational blocks such as greetings and general social chat, aimed at establishing the robot’s social presence and gather user preferences through interaction.
        \item \textbf{Educational Elements}: Techniques designed to foster engagement and give "reading tips",  such as brief co-reading or creating a bookmark with a reading plan.
    \end{itemize}
 \textbf{2. LLM-Generated Content}:
To achieve scalability and personalization, certain dialogue components were designed with rule-based structures but included dynamic placeholders that were populated offline using LLM-generated content (see Figure \ref{fig:figure_1},  blue blocks in the center).
    \begin{itemize}
        \item \textbf{Personalized Book-Related Dialogue Blocks}: The dynamic utterances (to fill in the placeholders) were LLM-generated and personalized for child’s profile (e.g., interests, reading preferences) and specific book content (for the LLM prompt high level structure see Figure \ref{fig:figure_1}, right-hand section). For instance, if the child’s user model indicated an interest in adventure, the robot might ask, “If you could fly like Peter Pan, where would you go first?”

    \end{itemize}
\textbf{3. Human-LLM Collaborative Evaluation as Safety Mechanism}:
To ensure that all interactions met quality and safety standards, a validation pipeline was implemented:
    \begin{itemize}
        \item \textbf{LLM-Based Evaluator}:  The use of LLM-based evaluation tools for natural language generation (NLG) represents a promising advancement for assessing the outputs of other prompts. Among the available methods, directly prompting LLMs to provide scores or judgments on generated content has proven effective and demonstrated strong correlations with human judgments \cite{gao_llm-based_2024}. We implemented a prompt-based validator that evaluated LLM outputs using the same criteria provided to the original generation prompt. The metrics included appropriateness, understandability, accuracy, relevance, engagement potential, and reflectiveness and assigning scores to each criterion. 
        \item \textbf{Human Moderation}: A final layer of human validation ensured developmental suitability and alignment with the project’s educational goals before the content was incorporated into interactions. This step involved evaluating, refining or regenerating content to fit the educational context. Moderators primarily focused on utterances with lower scores, while high-scoring content was only glanced through, allowing for targeted attention where it was most needed.
    \end{itemize}
Though labor-intensive, this combined approach was significantly more efficient than requiring moderators to familiarize themselves with 100 books and user profiles and validating over 2,000 utterances. The LLM-based evaluator allowed moderators to focus on targeted refinements. This approach enabled both the scale and the level of personalization required while maintaining the quality and appropriateness of the contents. 

Consider an example where child is assigned the book \textit{Peter Pan}, and their user model includes "favorite motifs in a book = pirates." As part of the pipeline, the LLM is prompted to generate various utterances for the book-related dialogue, including a humor-based utterance, which results in: \textit{"Why don’t pirates ever use grammar checkers? Because they’re always speaking 'Arrrr-chaic English!'"}. This utterance is forwarded to the LLM-based validator, which evaluates its suitability. The validator flags the joke as overly complex for the target age group and assigns a low score. Based on this feedback, a human moderator triggers content regeneration, obtaining in a more viable joke.

An example of a resulting personalized dialogue, tailored to incorporate the child’s interests, the specific content of their assigned book, and structured according to project’s educational goals.:
\begin{quote}
Robot: ...so when Peter flew to Neverland, I thought of you, Jip.  
I know you’d love to fly like him [User model/memory-based personalization].  What did you think of that part? [Question object] 

Jip: Oh, I loved it! It was so exciting and made me wish I could fly too.  

Robot: I know, right? [Pre-generated response object]
Flying would make every adventure so much more thrilling. [Opinion/observation object] 
If you could fly, what would be the first place you'd explore? [Follow-up question] 

Jip: I think I’d fly over a jungle and try to find hidden treasure!  

Robot: That’s a perfect adventurer’s choice, Jip! [Pre-generated response object].
Speaking of treasure, what did you think about the clever tricks Peter used to outsmart Captain Hook? [Follow-up question object] 

Jip: It was brilliant! I didn’t expect him to outsmart Hook like that.  

Robot: Exactly! Peter’s quick thinking always saves the day. Your adventurous spirit reminds me of a pirate joke. Want to hear it?  

Jip: Yes, please!
Robot: What do you call a pirate who likes to skip school? Captain Hooky! [\textit{humor object}]

\end{quote}

While a comprehensive analysis of our results is still underway, preliminary observations suggest that our hybrid framework successfully delivered engaging, personalized, purposeful, and age-appropriate child-robot interactions. Children found the dialogues interesting, helpful, and thought-provoking, expressing enthusiasm about continuing book-related conversations with a robot in the future. 

Although LLM was only provided with high-level metadata about each book and very limited information about the child, it was able to leverage information effectively and generate a relevant and engaging dialogue. Children reported that the dialogues felt directly aligned with them personally, ultimately deepening their overall engagement and creating the feeling of being “known by robot Leo”.

Finally, the multi-layered validation pipeline proved to be an effective approach for quality assurance. The LLM-based evaluator we implemented, while very basic, already proved to be "self-critical" and effective in assessing the quality of generated content, even humor.

\section{Future Work}

Guided by the lessons from the Robot-Bookworm project, we aim to advance the hybrid framework for dialogic learning in child-robot interactions. Our future work will focus on advancing personalization, adaptiveness, and scalability while maintaining safety and pedagogical alignment.

A key avenue for improvement involves the integration of Knowledge Graphs (KGs) and Retrieval-Augmented Generation (RAG) techniques into the system. KGs can provide structured, domain-specific knowledge, and enriched child profiles, offering a more reliable base for personalization. RAG, on the other hand, can enable the system to retrieve contextually relevant, factually grounded information dynamically, improving the quality and accuracy of generated dialogues. Together, these technologies promise to enhance both personalization and factual integrity in educational interactions.

Expanding the system's adaptive capacities without compromising safety is another priority. Incorporating real-time feedback mechanisms could enable the system to dynamically adjust dialogues based on children's immediate responses. However, to mitigate risks, our current approach emphasizes offline content generation and quality control before involving children in interactions. This ensures that all dialogue components are thoroughly validated for safety and alignment with educational goals. Future iterations may cautiously explore integrating limited real-time adaptability in highly controlled environments as technologies for real-time validation and dynamic filtering advance.

To improve scalability while maintaining quality and safety, we will enhance the automated evaluator by integrating advanced tools such as Natural Language Understanding (NLU) to ensure coherence and contextual relevance in generated dialogues, child-friendly language filters to guarantee age-appropriate and inclusive language, and fact-checking systems to verify the accuracy of generated content. These enhancements will strengthen the validation pipeline, reducing reliance on human moderation and allowing human evaluators to focus on refining and approving outputs flagged by automated systems. 

\section{Acknowledgments}

\bigskip
\noindent We sincerely thank the children and teachers of the participating schools for their invaluable contributions to this research. This work was supported by the Gravitation Programme Hybrid Intelligence, funded by the Nederlandse Organisatie voor Wetenschappelijk Onderzoek (NWO). Grant No. 024.004.022.

\bibliography{references}
\end{document}